# Deep Reinforcement Learning for Stock Portfolio Optimization

Le Trung Hieu

*Abstract*—**Stock portfolio optimization is the process of constant re-distribution of money to a pool of various stocks. In this paper, we will formulate the problem such that we can apply Reinforcement Learning for the task properly. To maintain a realistic assumption about the market, we will incorporate transaction cost and risk factor into the state as well. On top of that, we will apply various state-of-the-art Deep Reinforcement Learning algorithms for comparison. Since the action space is continuous, the realistic formulation were tested under a family of state-of-the-art continuous policy gradients algorithms: Deep Deterministic Policy Gradient (DDPG), Generalized Deterministic Policy Gradient (GDPG) and Proximal Policy Optimization (PPO), where the former two perform much better than the last one. Next, we will present the end-to-end solution for the task with Minimum Variance Portfolio Theory for stock subset selection, and Wavelet Transform for extracting multi-frequency data pattern. Observations and hypothesis were discussed about the results, as well as possible future research directions.**

## I. Introduction

In this project, we will explore the task of stock trading using reinforcement learning. To be specific, we will work on the task of portfolio optimization, where the stock weight distribution of the portfolio will be adjusted at the beginning of each day to maximize profits and constraining some certain risks [1].

The current main applications of machine learning to stock trading is through a price prediction network of the next market price state. As a supervised regression learning problem, this idea is straightforward to implement. Unfortunately, the network prediction is not equal to the actions that the trading agents should take. Translating from price prediction to agent action usually involves hard-coded logic layer, which is not extensible and generalized. Therefore, reinforcement learning was applied to utilize the price prediction model for the trading agents to devise optimal action plans.

The first wave of research on applying reinforcement learning to financial market dates back to 1997 [2]. There are existing works on portfolio management using reinforcement learning [3]. However, they test it on crypto-currency market which might not generalize well to stock market since crypto-currency is more volatile and stochastic, and the strong overall increasing trend given the recent hype. Besides, they test it only on the baseline of Deep Reinforcement Learning. So, we will extend the work to stock market which is fluctuating in both directions in contrast with the increasing trend of cryptocurrency. Besides that, in contrast with the basic assumptions about stock market [4]-[6], we will incorporate transaction cost and risk factor into our state and reward system as well. We will explore different state-of-the-art schemes of Deep Reinforcement Learning for this task.

Next, we use Minimum Variance Portfolio Theory to select a subset of stocks to construct the portfolio, since we can get a lower-risk portfolio. If we choose all the stocks to construct the portfolio, our portfolio will rely heavily on the overall market trend, so it is hard to make profit in bear market. We also perform Price Data Denoising using Wavelet Transform, so our agent can exploit both high-frequency patterns in original data(since it has all the noise from high-frequency trading) and low-frequency patterns in denoised data(since it removes the noise and uncover the underlying low-frequency pattern).

After that, we discuss about the algorithms we implement for this task. First, we go through the common Deep Deterministic Policy Gradient (DDPG) used by many existing works. Then, we discuss about two new variants of DDPG: GDPG and PPO. Finally, we reach to our results and observations. The overview pseudo-code of each deep reinforcement learning implementation is attached in the appendix.

Our main contributions in this paper are summarized as:

- Extend the work to stock market which is more realistic than cryptocurrency market, and propose a better problem formulation for more realistic simulations.

-Explore newer state-of-the-art deep reinforcement algorithms for the task.

- Better end-to-end optimization with Minimum Variance Portfolio Theory and Price Data Denoising using Wavelet Transform.

## II. Problem Formulation

Given a period of time, for example one year, the investor will invest in a portfolio of stocks. To decrease the portfolio risk, as commonly done, we maintain a portfolio of $m$+1 assets, with 1 risk-free asset (Money) and $m$ risky stock assets.

After we train the agent, we will do back-testing on test dataset to assess its performance. To ease the purpose of back-testing, here are two hypotheses about the market that we assume. Note that these assumptions are realistic given market with high volume transactions:

1) **Zero Slippage:** The market's liquidity is high enough that a **trade** can be transacted at exactly the price when order is placed.

Le Trung Hieu is with National University of Singapore, Singapore (e-mail: e0072405@u.nus.edu).

2) **Zero Market Impact:** The investment by the agent is insignificant to not affect the market at all.

Here is the process for updating a portfolio daily. The portfolio at the beginning of the day will change during the day, due to the price fluctuation of each individual stock. At the end of the day, we will reallocate the weights of each stock to get a new profit, which result in a portfolio that remains the same until next day market open. The same process happens again. Note that we assume the closing price of the current day is equal to the open price of the next day, which we believe is reasonable.

We can see that the action now are actually the portfolio weights. Therefore, this task is actually a continuous action space reinforcement learning task. Next, we are going to define the states, actions and rewards of this agent. Before that, we will go through some terminologies:

1) Price Vector: $v_t$ of period *t* stores the closing price of all assets in period t: $v_{\{i,t\}}$ is the closing price of asset I in period *t*. Note that the price of first asset is always constant since it is risk-free cash.

$$v_t = (v, v_{1,t}, v_{2,t}, .., v_{m,t}) \quad (1)$$

2) Price **Relative** Vector: $y_t$ is defined as the element-wise division of $v_t$ and $v_{t-1}$

$$y_t = \left(1, \frac{v_{1,t}}{v_{1,t-1}}, \frac{v_{2,t}}{v_{2,t-1}}, \dots, \frac{v_{m,t}}{v_{m,t-1}}\right)^T \quad (2)$$

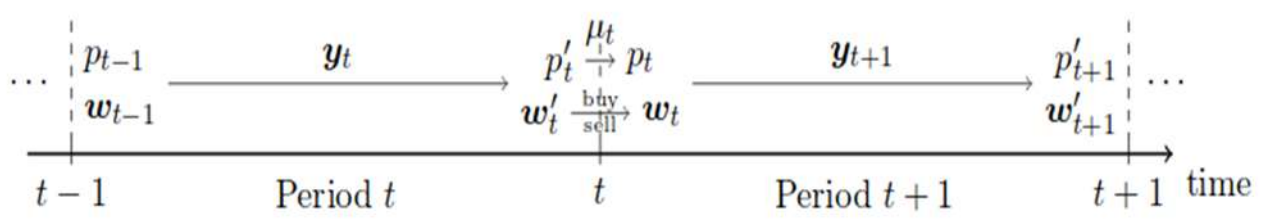

Fig. 1. During day *t*, market movement (represented by Price Relative Vector $y_t$) transforms the portfolio weights and portfolio values from $w_{t-1}$ and $p_{t-1}$ to $w'_t$ and $p'_t$. Then, at the end of day, we adjust the portfolio weights from $w'_t$ to $w_t$, which incurs transaction cost and shrinks the portfolio from $p'_t$ to $p_t$.

3) Portfolio weights and values after market movement:

$$w_t' = \frac{y_t \odot w_{t-1}}{y_t \cdot w_{t-1}} \quad (3)$$

where $\odot$ is element-wise multiplication, and $\cdot$ is dot product between two column vectors $y_t$ and $w_{t-1}$

$$p'_t = (y_t \cdot w_{t-1}) p_t \quad (4)$$

## A. State

State stores the history of prices of each stock in the portfolio over a window of time.

Therefore, the shape of the state will be (batch size, number of assets, window size, number of features).

## B. Action

Action now will become the weight distribution of the portfolio at each period's end, after the effect of market movement during the day. Therefore, the action space will be continuous and will need to use continuous action space policy gradient to tackle this task.

## C. Reward

A simple scheme of reward of each action is the change in the portfolio value during the market movement. However, this reward is not realistic because it is missing two important factors. Firstly, it lacks the transaction cost incurred with re-allocating portfolio at the end of day. Furthermore, the reward does not take into account the risk or volatility factor of the asset. We will encode this information inside our reward function.

We observe that for normal Markov Decision Process, the reward takes a form of discounted sum of rewards $\sum_{t=1}^{T} \gamma^t r(s_t, a_t)$, however for the case of portfolio management, the next wealth at period *t* actually depends on the wealth at period *t*−1 and the reward *r* in the form of product instead of sum: $New_Wealth = Old_Wealth * Reward$. Therefore, a slight modification of taking the logarithm of reward is used to transform the product form to normal summation form.

Therefore: Reward = log(wealth change - transaction cost) + (sharpe ratio that represents volatility factor).

$$r(s_{t-1}, w_{t-1}) = log\left(y_t \cdot w_{t-1} - \mu \sum_{i=1}^{M} |w_{i,t-1} - w'_{i,t-1}|\right) + \beta A \quad (5)$$

where

$$A = \sum_{i=1}^{M} w_{i,t-1} \frac{\frac{v_{i,t-1} - v_{i,t-L}}{v_{i,t-L}}}{std\left(\left[\frac{v_{i,t-1} - v_{i,t-2}}{v_{i,t-2}}, \dots, \frac{v_{i,t-L+1} - v_{i,t-L}}{v_{i,t-L}}\right]\right)}$$

**Preprocessing**

## D. Stock Selection for Portfolio

To reduce the vast search space of the portfolio state, we will reduce the number of stocks in a portfolio. We will find a minimum variance portfolio of 6 stocks from the overall stocks list [7]. The empirical covariance for each pair of stocks is obtained using historical data from the training set. For every combination of 6 out of 50 stocks, we compute its optimal weight

$$w^* = \frac{C^{-1}1}{1^T C^{-1} 1}$$

that produces the minimal variance

$$\sigma^2 = \frac{1}{1^T C^{-1} 1}$$

## E. Data Denoising

The time series data of stock usually oscillates frequently. To understand this, we may consider two kinds of trading participants: one is to take rational actions of buying or selling, and this is represented by main tendency of the data. The other one is to take random actions since they may have other consideration (e.g. need to get money for emergency usage), and this is represented by oscillations (noise) in data. Denoising is necessary to help us understand the rational strategy, and then develop good policy [8].

We use the discrete wavelet transform to denoise on 1-D data, because it is applicable to non-stationary series [9], which means the frequency could change in time. Wavelet Transform has been frequently applied to the financial market as well [10], [11]. It firstly decomposes the original data to generate the approximation and detail coefficients. The approximation coefficients approximate the tendency of the data with less oscillation, and the detail coefficients provide the frequency of the oscillation based on the approximation. Fig. 2 shows an example of two coefficients generated from the original data.

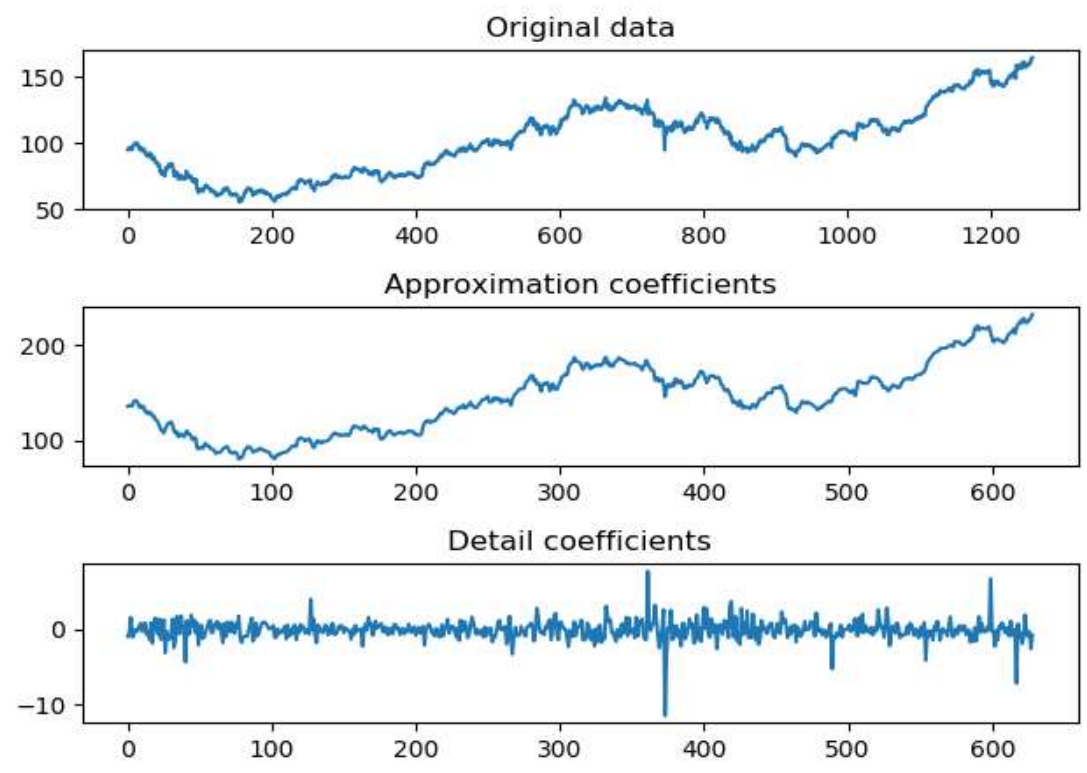

Fig. 2. Original data and coefficients.

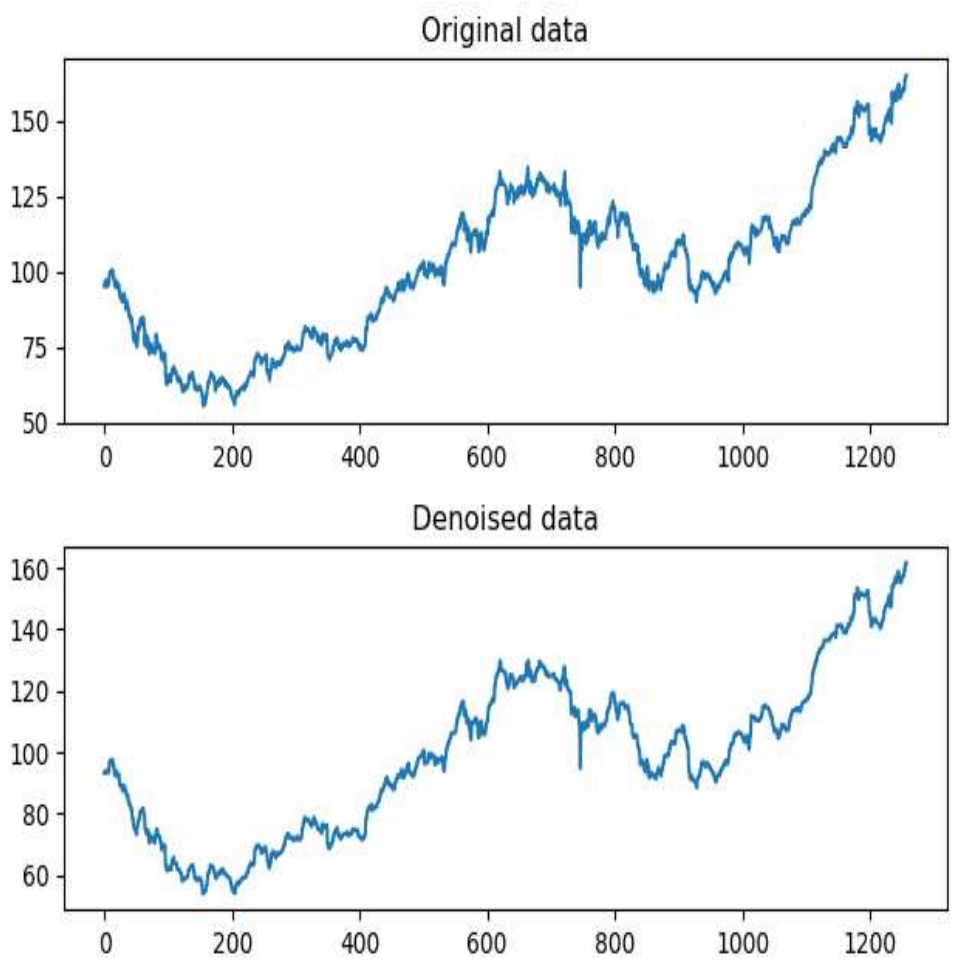

Fig. 3. Original data and denoised data.

This decomposition is reversable, meaning that we can reconstruct the original data by these coefficients. To denoise, we should remove some of the detail coefficients. Therefore, we use a threshold $T$ to filter out small noise. Using the formula below adapted from [12]:

$$T = \frac{\sqrt{2\ln N} * median(|D|)}{0.6745} \tag{6}$$

where $N$ is the size of original data, and D is the detail coefficients.

Next, for each d in detail coefficients, apply:

$$\begin{aligned} d &\leftarrow 0 \qquad |d| \leq T \\ d &\leftarrow d - T |d| > T \end{aligned} \tag{7}$$

Then when we reconstruct by filtered detail coefficient, the result will tend to approximation and less oscillate because zero in detail coefficient prevent oscillating based on approximation data. Fig. 3 using the same example as Fig. 2 above shows the effect of denoising.

## III. Methods

### A. Deep Deterministic Policy Gradient

To tackle this problem, we need a Reinforcement Learning paradigm that can deal with continuous action space. Recall that Deep-Q Learning will take in a state **s** and return a vector $a = [a_1, a_2, \dots, a_A]$where $a_i$ represents the probability of action *i*. Naively extending this scheme to continuous action space means to extend the size of vector *a* to a very large number, which will not work well.

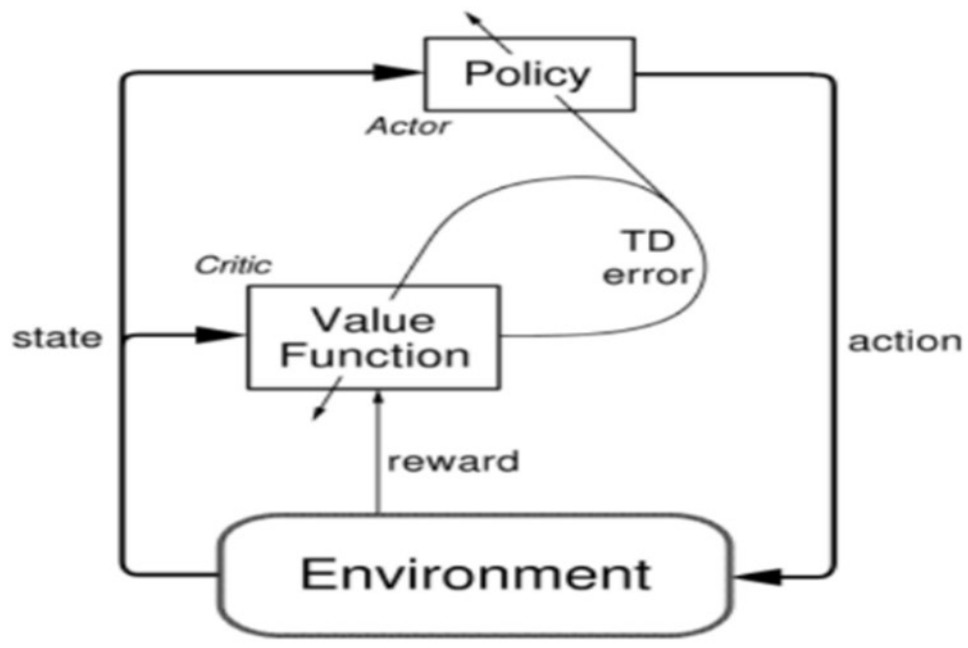

Fig. 4. Actor-critic architecture.

DDPG solves this issue by following the actor-critic architecture in Fig. 4. An actor is used to output a vector that represents the expected action, which can be seen as a policy gradient method. Next, a critic is used to evaluate the effectiveness of the output of the actor, and output Q-value that measures the efficiency of the actor. Critic loss can be used to update back the actor.

DDPG is based on the actor-critic architecture, with some modifications. Firstly, the actor and critic networks are approximated using a Deep Neural Network ($\theta^{\mu}$ for actor and $\theta^{Q}$ for critic). Next, we use the idea of separate target network for both actors and critics, similar to Deep-Q Learning, in order to stabilize the learning. The networks are also randomly noised as a scheme to balance the exploration-exploitation issue in Reinforcement Learning. More information can be found in [13].

So the question now is how to update the actor policy. We need to calculate the gradient of policy loss with respect to actor parameters $\nabla_{\theta^{\mu}} J$ . According to the Deterministic Policy Gradient Theorem in the original paper:

$$\begin{aligned} \nabla_{\theta^{\mu}} J &\approx E_{s_t \sim \rho^{\beta}}\left[\nabla_{\theta^{\mu}} Q(s, a|\theta^{Q})|_{s=s_t, a=\mu(s_t|\theta^{\mu})}\right] \\ &= E_{s_t \sim \rho^{\beta}}\left[\nabla_a Q(s, a|\theta^{Q})|_{s=s_t, a=\mu(s_t)} \nabla_{\theta^{\mu}} \mu(s|\theta^{\mu})|_{s=s_t}\right] \end{aligned}$$

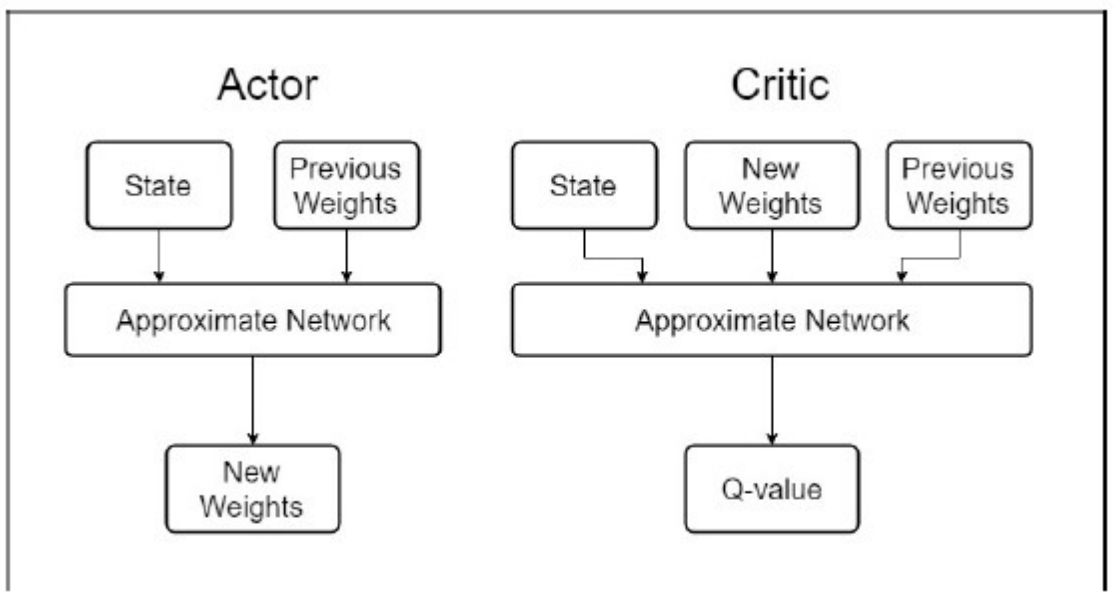

Fig. 5. DDPG actor-critic architecture. Note that actor and critic deep neural networks take in both current state and previous portfolio weights. This is because it needs to learn not to diverge too much from the previous weights to prevent high transaction cost.

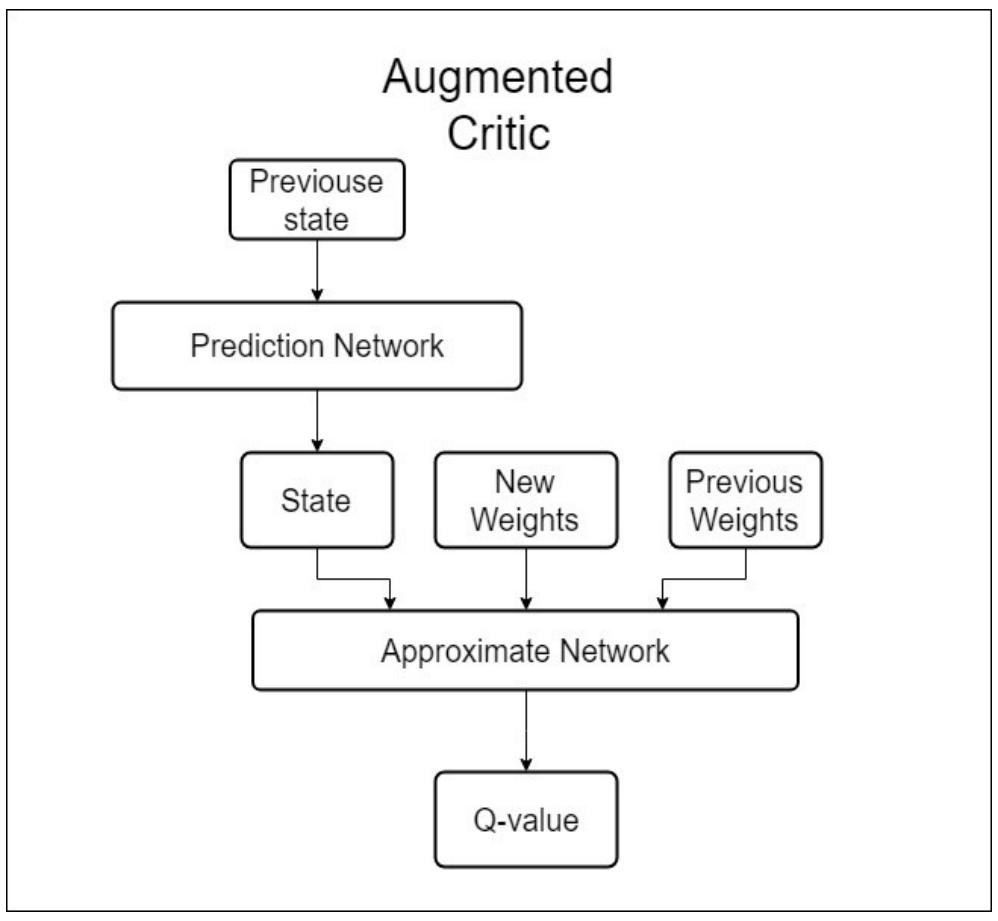

Fig. 6. GDPD augmented critic network.

## B. Generalized Deterministic Policy Gradient:

One of the problems with DDPG is that it assumes stochastic state transition. In fact, for most planning problems such as autonomous vehicle, the state transition might be a combination of both stochastic state transition (when in dynamics) and deterministic state transition (when noises are weak). However, the gradient of DDPG in such a new assumption is not well-defined, and can lead to weird behavior frequently. Main problem is that the model-free DDPG is known to have high sampling complexity, which makes learning difficult. Transforming DDPG into completely model-based can reduce the sampling complexity. Unfortunately, purely model-based reinforcement learning can lead to slow convergence rate(or sometimes huge divergence) if the environment is highly dynamic (which is especially true for stock market).

An idea is to combine model-free and model-based approaches in some meaningful ways. With the above mentioned insights, a new variation of DDPG is proposed, which is called General Deterministic Policy Gradient [14]. GDPG intuition is to maximize the long-term reward of the augmented MDP(which is approximated by a model-based network) to reduce sample complexity, but at the same time constrain it to be less than the long-term reward of the original model-free MDP:

$$\max_{\theta} J_*(\theta^\mu), \qquad \text{s.t} \quad J_*(\theta^\mu) \le J(\theta^\mu)$$

Using Lagrangian dual theorem, the new objective function is transformed into:

$$\min_{\alpha \ge 0} \max_{\theta} J_*(\theta^\mu) + \alpha\big(J(\theta^\mu) - J_*(\theta^\mu)\big)$$

To update policy of actor, we will find gradient of $J_*(\theta^\mu) + \alpha\big(J(\theta^\mu) - J_*(\theta^\mu)\big)$

$$\nabla_{\theta^\mu} J(\theta^\mu) = \frac{1}{N}\sum_i (1-\alpha) * \nabla\theta^\mu \mu(s|\theta^\mu)\nabla_a Q_*(s,a|\theta^{Q_*}) + \alpha\nabla_{\theta^\mu}\mu(s|\theta^\mu)\nabla_a Q(s,a|\theta^Q) \qquad (8)$$

The main difference between DDPG and GDPG is that GDPG maintains a prediction neural network model, which can predict the next market state given the current state. This prediction neural network is used to build an augmented critic network as in Figure 6. The actor is updated based on a combination of gradients from both original model-free critic network and augmented model-based critic network.

## C. Proximal Policy Optimization

Proximal Policy Optimization (PPO) is another variant of DDPG, which aims to improve updating actor policy.

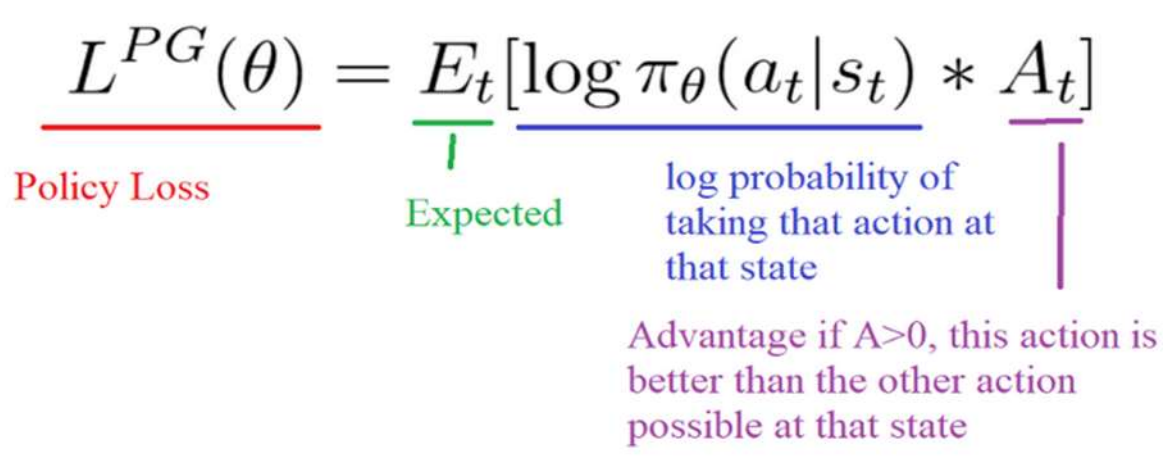

Fig. 7. Policy loss function.

Recall the original policy gradient objective function in Fig. 7, we find that it is appealing to perform multiple steps of optimization on this loss using the same trajectory. Doing so is not well-justified, and empirically it often leads to destructively large policy updates [14].

To tackle this problem, PPO was proposed to make use of a surrogate objective function of the original policy loss function. Instead of using log to trace impact of action, we will use **the ratio between the probability of action under current policy divided by the probability of the action under previous policy.** The ratio is formally defined as:

$$r_t(\theta) = \frac{\pi_\theta(a_t|s_t)}{\pi_{\theta_{old}}(a_t|s_t)}$$

With this new definition, the objective function now becomes:

$$L^{CPI}(\theta) = \hat{E}_t\left[\frac{\pi_\theta(a_t|s_t)}{\pi_{\theta_{old}}(a_t|s_t)}\widehat{A_t}\right] = E_t\big[r_t(\theta)\widehat{A_t}\big]$$

However, without any constraint, this policy still leads to excessive large policy updates. PPO proposed to clip the objective function to penalize changes in policy that lead ratio $r_t(\theta)$ far away from 1. Therefore, the ratio is clipped to the range of $[1-\varepsilon, 1+\varepsilon]$. This net surrogate objective function can constrain the update step in a much simpler manner, and experiments in the PPO paper show it does outperform the original objective function in terms of sample complexity.

(Note that $A$ is the advantage value, which is defined as $A_\pi(s,a) = Q_\pi(s,a) - V_\pi(s)$, which shows how good an action is compared to average of other actions at that state. In PPO, we will estimate the advantage value as $\sum_{t^{'}>t}\gamma^{t^{'}-t}r_{t^{'}} - V(s_t)$.

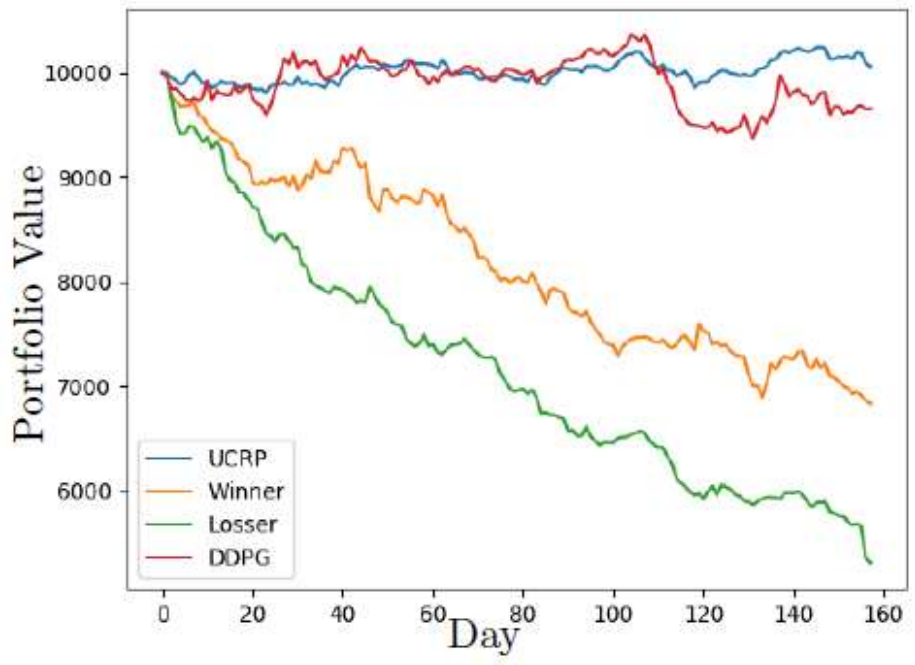

Fig. 8. Result on safe portfolio with *K*=6.

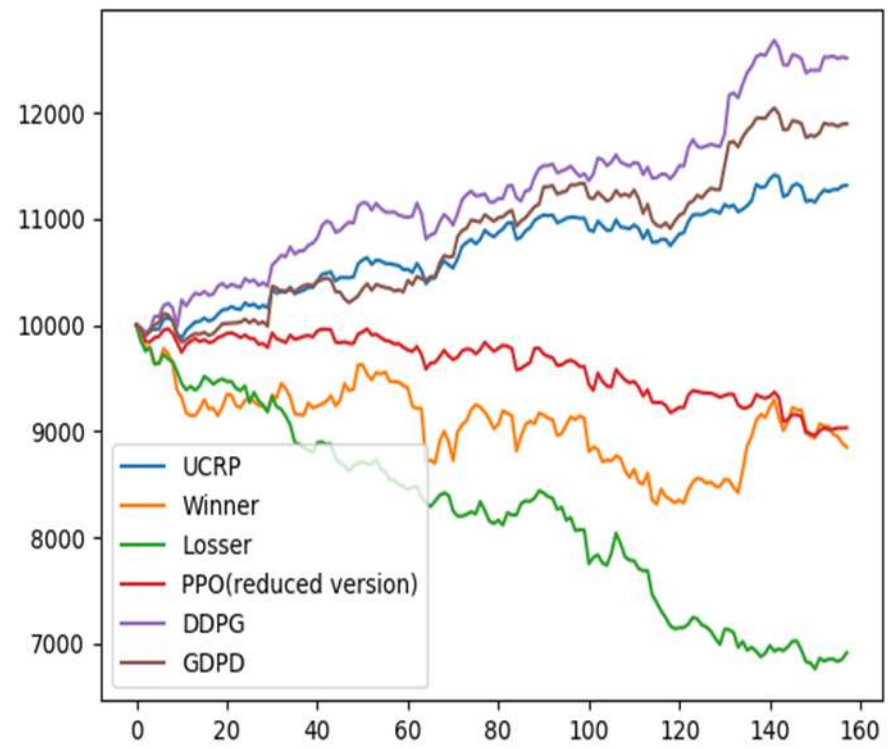

Fig. 9. Final result.

## IV. Results

We experimented with stocks in the training dataset from 01/09/2012 to 31/12/2016. Then we back-test our agents from 01/01/2017 to 01/09/2017. We make use of 3 features (close price, high price, close price after wavelet transform).

The network we use for actor and critic are Convolutional Neural Networks. The neural network used for model the state-transition of GDPG is a Long Short Term Memory Network. Since the focus of this project is on Reinforcement Learning, we will not go into details of these networks.

**Baselines:** To compare with DDPG, GDPG and PPO, we use 3 baselines:

[(a)]

1) **Uniform constant rebalanced portfolios (Benchmark):** At the end of each day, the portfolio is adjusted such that the weights are same for all stocks. This is the common benchmark used for portfolio management research.

2) **Follow-the-winner:** Shift all the portfolio weights to the stock that has the highest return yesterday. This is based on the belief that it will keep continuing today.

3) **Follow-the-loser:** Shift all the portfolio weights to the stock that has the lowest return yesterday. This is based on the belief that it has highest chance to improve today.

Firstly, we construct our portfolio with the K stocks with minimum variance portfolio among all possible combination of K stocks, as described in section 3.1. Unfortunately, the result seems not promising, as in Fig. 8. Our hypothesis is that choosing the combinations of stocks with lowest risk results in a lower-risk portfolio, but it also means the promising profits cannot be high as well.

Instead, next we choose a portfolio of "AAPL", "PG", "BSAC", "XOM" from different industries to slightly diversify the portfolio. The result is illustrated in Fig. 9.

### A. Observations

1) The **best** stock selection for initial portfolio, as presented in section 3.1, is not a good idea. It gives a too low-risk portfolio with also very low potential profits.

2) Follow-the-winner and Follow-the-loser performs poorly due to transaction cost incurred. That's why existing works do not take into account transaction costs will lead to very misleading results.

3) DDPG has the best performance. However, in theory GDPG can reach a better performance than DDPG by reducing **sample** complexity. Our hypothesis for this discrepancy is that since GDPG is very sensitive to the accuracy of the model-based state-transition modelling [14], in this case we use LSTM which is not a very good model. Therefore, next price state prediction model is a very important component for GDPG performance, and it is a potential future work to explore how price prediction model accuracy affects GDPG.

4) PPO performs **much worse** compared to other agents, despite the promising mathematical characteristics. Our observation is that if the PPO actor network takes in the previous portfolio weights, it will reach a similar path as UCRP. If we remove the previous portfolio weights from actor network, it will reach the path as in Figure 9. The reason still remains unclear to us, it could be because stock market might not be suitable for PPO.

### B. Conclusion and Future Work

In this project, we have explored the task of portfolio management with reinforcement learning, and obtained some insights from the result. There are many future directions to continue from here. For example, we can include all stocks in the portfolio, and the agent can learn to put 0 in many stocks except a few stocks. However, this task is easy to get stuck in local minimum, and transaction cost prevents large shift of weights between days. Another direction is to provide better networks for actor, critic and **especially the state transition network of GPDG**.

## Conflict of Interest

The authors declare no conflict of interest.

## Author Contributions

Author Le Trung Hieu is the sole author of this paper.

## References

[1] R. A. Haugen and R. A. Haugen, *Modern Investment Theory*, vol. 5, Prentice Hall Upper Saddle River, NJ, 2001.

[2] J. Moody, L. Z. Wu, Y. S. Liao, and M. Saffell, "Performance functions and reinforcement learning for trading systems and portfolios," *Journal of Forecasting*, vol. 17, pp. 441–470, 1998.
[3] Z. Y. Jiang, D. X. Xu, and J. J. Liang, "A deep reinforcement learning framework for the financial portfolio management problem," 2017.
[4] J. Carapuco, R. Neves, and N. Horta, "Reinforcement learning applied to forex trading," *Applied Soft Computing*, vol. 7, pp. 783–794, 2018.
[5] G. Jeong and H. Y. Kim, "Improving financial trading decisions using deep q-learning: Predicting the number of shares, action strategies, and transfer learning," *Expert Systems with Applications*, vol. 117, pp. 125–138, 2019.
[6] J. Zhang and D. Maringer, "Indicator selection for daily equity trading with recurrent reinforcement learning," in *Proc. the 15th Annual Conference Companion on Genetic and Evolutionary Computation*, 2013, pp. 1757–1758.
[7] R. Clarke, H. D. Silva, and S. Thorley, "Minimum-variance portfolio composition," *The Journal of Portfolio Management*, vol. 37, no. 2, pp. 31–45, 2011.
[8] K. K. Lai and J. Huang, *The Application of Wavelet Transform in Stock Market*, JAIST Press, 2007.
[9] M. Rhif, A. B. Abbes, I. R. Farah, B. Martınez, and Y. F. Sang, "Wavelet transform application for/in non-stationary time-series analysis: A review," *Applied Sciences*, vol. 9, no. 7, pp. 1345, 2019.
[10] Z. X. Liu, "Analysis of financial fluctuation based on wavelet transform," Francis Academic Press, 2019.
[11] J. Nobre and R. F. Neves, "Combining principal component analysis, discrete wavelet transforms and xgboost to trade in the financial markets," *Expert Systems with Applications*, vol. 125, pp. 181–194, 2019.
[12] D. B. Percival and A. T. Walden, "Wavelet-based signal estimation," *Cambridge Series in Statistical and Probabilistic Mathematics*, Cambridge University Press, pp. 393–456, 2000.
[13] T. P. Lillicrap, J. J. Hunt, and A. Pritzel, N. M. O. Heess, T. Erez, Y. Tassa, D. Silver, and D. P. Wierstra, "Continuous control with deep reinforcement learning," US Patent, 2017.
[14] Q. P. Cai, L. Pan, and P. Z. Tang, "Generalized deterministic policy gradient algorithms," 2018.
[15] J. Schulman, F. Wolski, P. Dhariwal, A. Radford, and O. Klimov. "Proximal policy optimization algorithms," 2017.

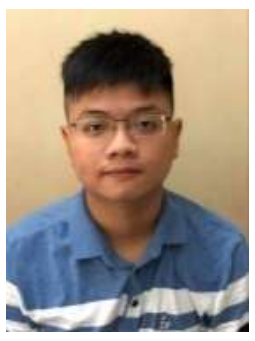

**Le Trung Hieu** was born in Vietnam, 1998. He is in his final year of honors bachelor degree at National University of Singapore, majoring in computer science. He is pursuing the artificial intelligence focus in his career and study path. His relevant experience includes a research attachment at NUS-Tsinghua lab, and computer vision engineer internships at Microsoft in Singapore and a startup in Israel. Before that, he worked in software engineering roles at Goldman Sachs and Sea Group in Singapore.